\documentclass[runningheads]{llncs}

\usepackage{eccv}

\usepackage{eccvabbrv}

\usepackage{graphicx}
\usepackage{booktabs}

\usepackage[accsupp]{axessibility}  

\usepackage{hyperref}

\usepackage{orcidlink}

\usepackage{epigraph}
\PassOptionsToPackage{prologue,dvipsnames}{xcolor}

\newcommand{\delete}[1]{}

\usepackage{color}
\usepackage {colortbl}
\definecolor{black}{rgb}{0,0,0}
\definecolor{blue}{rgb}{0,0,1}
\definecolor{red}{rgb}{1,0,0}
\definecolor{green}{rgb}{0,.5,0}
\definecolor{orange}{rgb}{0.75, 0.4, 0}
\definecolor{teal}{rgb}{0.0, 0.4, 0.4}
\definecolor{purple}{rgb}{0.65,0,0.65}
\definecolor{gray}{rgb}{0.94, 0.94, 0.94}

\newcommand{\renderer}{g}
\newcommand{\modelparams}{\psi}
\newcommand{\renderedimage}{x}
\newcommand{\camerapose}{\zeta}
\newcommand{\textprompt}{T}
\newcommand{\textembedding}{y}
\newcommand{\timestep}{t}
\newcommand{\noise}{\epsilon}
\newcommand{\noisepredictnet}{\phi} 
\newcommand{\bodypose}{\theta}
\newcommand{\bodyshape}{\beta}
\newcommand{\spatialpoint}{p}
\newcommand{\allpoints}{P}
\newcommand{\crossentropy}{CE}
\newcommand{\hyperparameter}{\eta}
\newcommand{\distancetoanchor}{d}

\begin{document}

\title{InterFusion: Text-Driven Generation of \\ 3D Human-Object Interaction}

\titlerunning{InterFusion}

\renewcommand{\thefootnote}{\fnsymbol{footnote}}
\footnotetext[1]{\label{corr}Corresponding authors: kevin.kai.xu@gmail.com; ruizhen.hu@gmail.com.}
\author{Sisi Dai\inst{1}
~
Wenhao Li\inst{1} 
~
Haowen Sun\inst{2} 
~
Haibin Huang\inst{3} 
~
Chongyang Ma\inst{3} 
~
Hui Huang\inst{2} 
~
Kai Xu\inst{1}$^{\ref{corr}}$
~
Ruizhen Hu\inst{2}$^{\ref{corr}}$}

\authorrunning{Dai et al.}

\institute{National University of Defense Technology \and 
Shenzhen University \and 
Kuaishou Technology \\
\url{https://sisidai.github.io/InterFusion/}}

\maketitle
\begin{figure*}
\centering
\includegraphics[width=\textwidth]{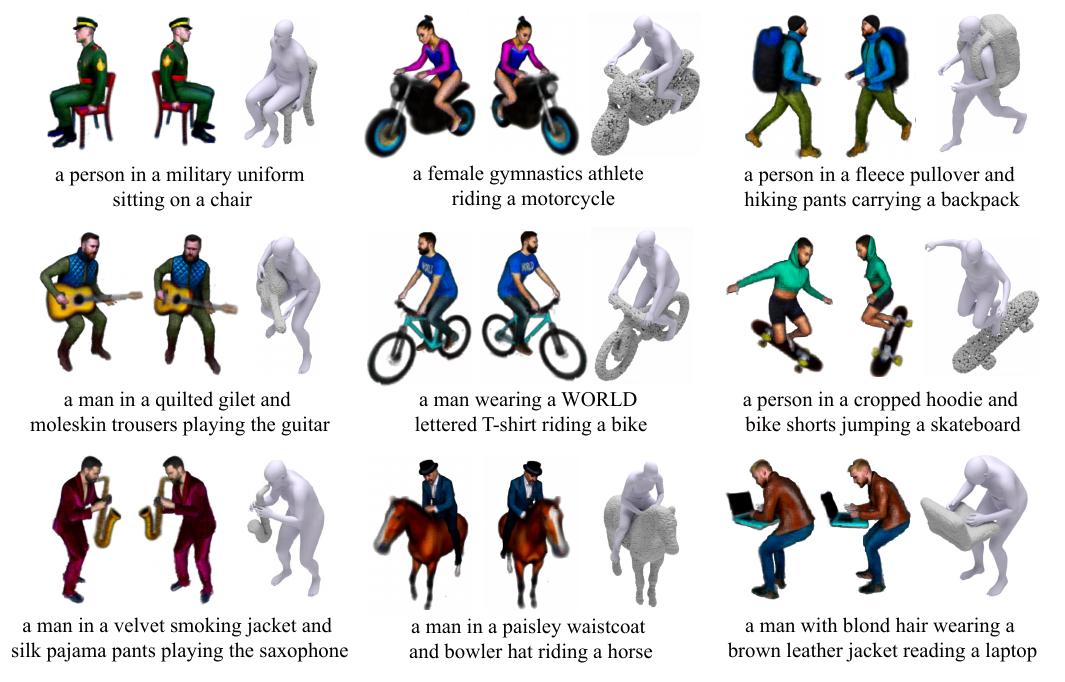}
\caption{Given a text prompt, our method InterFusion can generate diverse 3D scenes of a person interacting with an object.}
\label{fig:feature-graphic}
\end{figure*}

\begin{abstract}

In this study, we tackle the complex task of generating 3D human-object interactions (HOI) from textual descriptions in a zero-shot text-to-3D manner. We identify and address two key challenges: the unsatisfactory outcomes of direct text-to-3D methods in HOI, largely due to the lack of paired text-interaction data, and the inherent difficulties in simultaneously generating multiple concepts with complex spatial relationships. To effectively address these issues, we present InterFusion, a two-stage framework specifically designed for HOI generation. InterFusion involves human pose estimations derived from text as geometric priors, which simplifies the text-to-3D conversion process and introduces additional constraints for accurate object generation.  At the first stage, InterFusion extracts 3D human poses from a synthesized image dataset depicting a wide range of interactions, subsequently mapping these poses to interaction descriptions. The second stage of InterFusion capitalizes on the latest developments in text-to-3D generation, enabling the production of realistic and high-quality 3D HOI scenes. This is achieved through a local-global optimization process, where the generation of human body and object is optimized separately, and jointly refined with a global optimization of the entire scene, ensuring a seamless and contextually coherent integration. Our experimental results affirm that InterFusion significantly outperforms existing state-of-the-art methods in 3D HOI generation.


\end{abstract}

\section{Introduction}

The generation of 3D human-object interactions (HOI) stands as a critical challenge in the fields of computer vision and computer graphics, with far-reaching implications in virtual reality, augmented reality, animation, and embodied AI~\cite{puig2023habitat, sheridan2016human}. This task entails the creation of realistic 3D scenes where human figures interact with objects in ways that are not only physically plausible but also contextually relevant. 

Despite its potential, the field has faced significant obstacles, primarily due to the scarcity of large-scale interaction data. Traditional approaches have predominantly relied on motion capture (mocap) datasets or physics-based simulations for generating these interactions. Mocap datasets~\cite{ionescu2013human3, mahmood2019amass, von2018recovering} are limited by the specific scenarios they capture and are both costly and labor-intensive to produce.  These limitations have resulted in a notable gap in generating diverse and contextually rich HOI scenes, especially for novel or complex interactions. Conversely, recent advancements have introduced text-to-3D methods~\cite{poole2022dreamfusion, wang2022score}, marking a significant shift in the field. These methods harness the power of textual descriptions to generate 3D objects without direct 3D supervision, presenting a novel approach to 3D content creation.

In this study, we explore text-to-3D method in HOI task within a zero-shot manner, \ie, generating 3D scenes from textual descriptions using 2D diffusion models. Our key observations are two-fold: first of all, a direct application of text-to-3D method in HOI often leads to unsatisfied results like blurry textures and incorrect interactions, which is caused by the relative lack of paired text-interaction data during training and the difficulties of generating multiple concepts for diffusion-based methods~\cite{liu2022compositional}. Secondly, while collecting extensive interaction data is challenging, estimating human poses based on described interactions is more feasible. These pose estimations can serve as geometric priors in the HOI generation process. By integrating pose estimation, our method substantially simplifies the text-to-3D process, particularly in the geometry optimization stage for human body generation. It also provides additional constraints for object generation, ensuring accurate placement and alignment with the text's semantic content. More importantly, we can separate the generation of human body and object, and jointly refine the details with the estimated pose, allowing for a more coherent and detailed synthesis of the interaction scene.

Inspired by these observations, we introduce InterFusion, a novel two-stage framework designed for HOI generation. Specifically, in the first stage, rather than relying on precise 3D interaction data,  our approach instead collects a comprehensive dataset of  synthesized images that depict a wide range of interactions. From these images, we employ advanced 3D pose estimation technique~\cite{feng2021collaborative} to extract 3D human poses. Building upon these image-pose pairs, InterFusion develops a sophisticated codebook that establishes a mapping between interaction descriptions and 3D human poses, with the integration of the CLIP (Contrastive Language–Image Pretraining) embedding ~\cite{radford2021learning}. By leveraging these embeddings, our framework is able to interpret the nuances of interaction descriptions and translate them into accurate 3D pose representations. In the subsequent stage, InterFusion capitalizes on recent advancements in text-to-3D generation  ~\cite{poole2022dreamfusion}, as well as neural radiance fields ~\cite{mildenhall2021nerf}, using the estimated human poses to produce 3D HOI scenes with realistic appearances and high-quality geometry. This stage operates in a `local-global' manner. At the local level, the generation of the human body (SDS-H) and objects (SDS-O) is separately optimized, with the poses serving as additional constraints for the SDS. At the global level, the generation of the entire scene (SDS-I) is also guided by the integrated description and jointly optimized with SDS-H and SDS-O, ensuring a cohesive and contextually accurate representation of the HOI. Our experiments show that the quality of generation can be improved by a large margin and our approach outperforms state-of-the-art methods in HOI generation.   


To summarize, our contributions are as follows:

\begin{itemize}
\item We introduce a novel two-stage framework InterFusion, for zero-shot 3D human-object interaction generation from text, incorporating 3D pose estimation as geometry priors. 

\item  InterFusion leverages text-to-3D generation with a local-global optimization process. This strategy ensures seamless integration of human bodies and objects, producing realistic and high-quality 3D HOI scenes.

\item  InterFusion demonstrates significant improvements over existing methods in 3D HOI generation, showcasing its effectiveness in creating detailed, and contextually rich 3D interactions.
\end{itemize}

\section{Related Work}

\subsection{Human-Object Interaction Synthesis}
3D human-object interaction (HOI) generation is a challenging problem that has been studied widely by the computer vision and graphics community. Shape2Pose \cite{Kim:2014:SHS} generates plausible 3D human poses interacting with a given 3D object model by learning an affordance model from synthetic data. PiGraphs \cite{savva2016pigraphs} learns the distribution between human poses and object arrangements from a collected dataset with 3D scene scans and RGB-D videos, generating the interaction snapshots given action specifications and object models. Recently, the parametric human body models such as SMPL \cite{loper2015smpl, romero2022embodied, pavlakos2019expressive} are employed in interaction synthesis to overcome the lack of realism due to human body representations.
Benefiting from the PROX dataset \cite{hassan2019resolving} which consists of fitted SMPL models in captured 3D scenes, the more specific human-scene interaction has become an active research direction.
Given a 3D scene, PSI \cite{zhang2020generating} and PLACE \cite{zhang2020place} generate the 3D human body mesh represented as SMPL parameters through a conditional variational autoencoder \cite{kingma2013auto, sohn2015learning}.
POSA \cite{hassan2021populating} learns pose-specific priors to generate contacts conditioned by the given posed human, which can further guide the placement of the body mesh in a scene.
COINS \cite{zhao2022compositional} enables semantic control on interaction synthesis by embedding the action label together with the interacted object as the condition of the generative model.
More recently, fine-grained 3D interaction datasets~\cite{taheri2020grab,bhatnagar2022behave,fan2022articulated} are captured to promote this field. Relying on these datasets, some methods~\cite{diller2024cg, peng2023hoi, wu2024thor}, concurrent to this work, are proposed to explore text-guided 3D HOI generation. However, they remain constrained by distributions within the datasets.
While previous methods need ground truth 3D interaction data as supervision, our work, for the first time, attempts to break through the limitation of data requirement. We generate a wider range of realistic and detailed 3D HOI scenes, including both indoors and outdoors.

\subsection{Text-to-3D Content Synthesis}
Early methods \cite{chen2019text2shape,jahan2021semantics,liu2022towards} for text-to-3D shape generation require paired data of 3D data and the corresponding textual descriptions to learn the joint embedding space of shape and text for supervision, which limits their generality to unseen object categories.
Benefiting from large pre-trained text-to-image models and differentiable rendering techniques, breakthroughs in text-to-3D content generation have been achieved.
For example, DreamFields \cite{jain2022zero} and PureCLIPNeRF \cite{lee2022understanding} combine CLIP~\cite{radford2021learning} with neural radiance fields (NeRF) \cite{mildenhall2021nerf}, demonstrating the potential for zero-shot NeRF optimization. 
Meanwhile, CLIP-mesh \cite{mohammad2022clip} and Text2Mesh \cite{michel2022text2mesh} incorporate CLIP to optimize the 3D mesh representation, starting from an initial sphere mesh and an input base mesh, respectively.  
Recently, DreamFusion \cite{poole2022dreamfusion} and SJC \cite{wang2022score} enable NeRF optimization with guidance from pre-trained text-to-image diffusion models \cite{saharia2022photorealistic,rombach2022high} in place of CLIP, achieving more impressive results.
To improve DreamFusion, Magic3D \cite{lin2022magic3d} proposes a coarse-to-fine pipeline to generate the fine-grained mesh. TextMesh \cite{tsalicoglou2023textmesh} extends the geometry representation from NeRF to an SDF framework, thereby enhancing detailed mesh extraction and photorealistic rendering.
Among follow-up works, Latent-NeRF \cite{metzer2022latent} and Vox-E \cite{sella2023vox} utilize explicit 3D shapes to provide additional training signals for NeRF optimization. While Latent-NeRF utilizes a rough untextured object for shape sculpture, Vox-E takes multiview images of a fine-grained textured object to edit geometry and appearance. 
All the methods mentioned above focus on the generation or edition of a single subject, while ignoring the interaction between different subjects.

\subsection{Compositional Scene Generation}
Representing scenes as compositions of object representations facilitates enhanced controllability.
Numerous techniques incorporate additional information at the object level to perform compositional modeling of scenes, effectively separating object representations from the overall scene imagery.
For instance, some methods incorporate 2D semantic information such as segmentation labels \cite{zhi2021place}, instance masks \cite{yang2021learning, wu2022object}, or features from a pre-trained vision-language model \cite{mirzaei2022laterf}.
Some other methods \cite{guo2020object, ost2021neural, song2022towards} use 3D layout information by object-centric bounding boxes with canonical coordinates. When it comes to scene generation, \cite{nguyen2020blockgan, niemeyer2021giraffe, xu2023discoscene} use compositional representations to generate scenes in a controllable manner.
More recent approaches \cite{po2023compositional, lin2023componerf, cohen2023set} generate compositional 3D scenes from input text prompts. Different from existing methods, our HOI scene is generated automatically without the requirement of input layout. Moreover, the spatial relationship between human and object during interaction is much more complex and cannot be simply characterized using their bounding boxes.

\section{Our Method}

\subsection{Overview}
In this section, we formally introduce InterFusion with a focus on the zero-shot and text-driven generation of 3D human-object interactions (HOI). Specifically, the input is a triplet of text descriptions, $\textprompt=\{\textprompt^{H}, \textprompt^{O}, \textprompt^{I}\}$, specifying the desired human style, object style, and interaction type. The goal is to generate a detailed 3D scene, $\modelparams=\{\modelparams^{H}, \modelparams^{O}\}$, comprising a human model and an object model that not only adhere to the specified appearance styles but also exhibit a tailored spatial relationship to accurately reflect the described interaction. 

\begin{figure*}
    \centering
    \includegraphics[width=\textwidth]{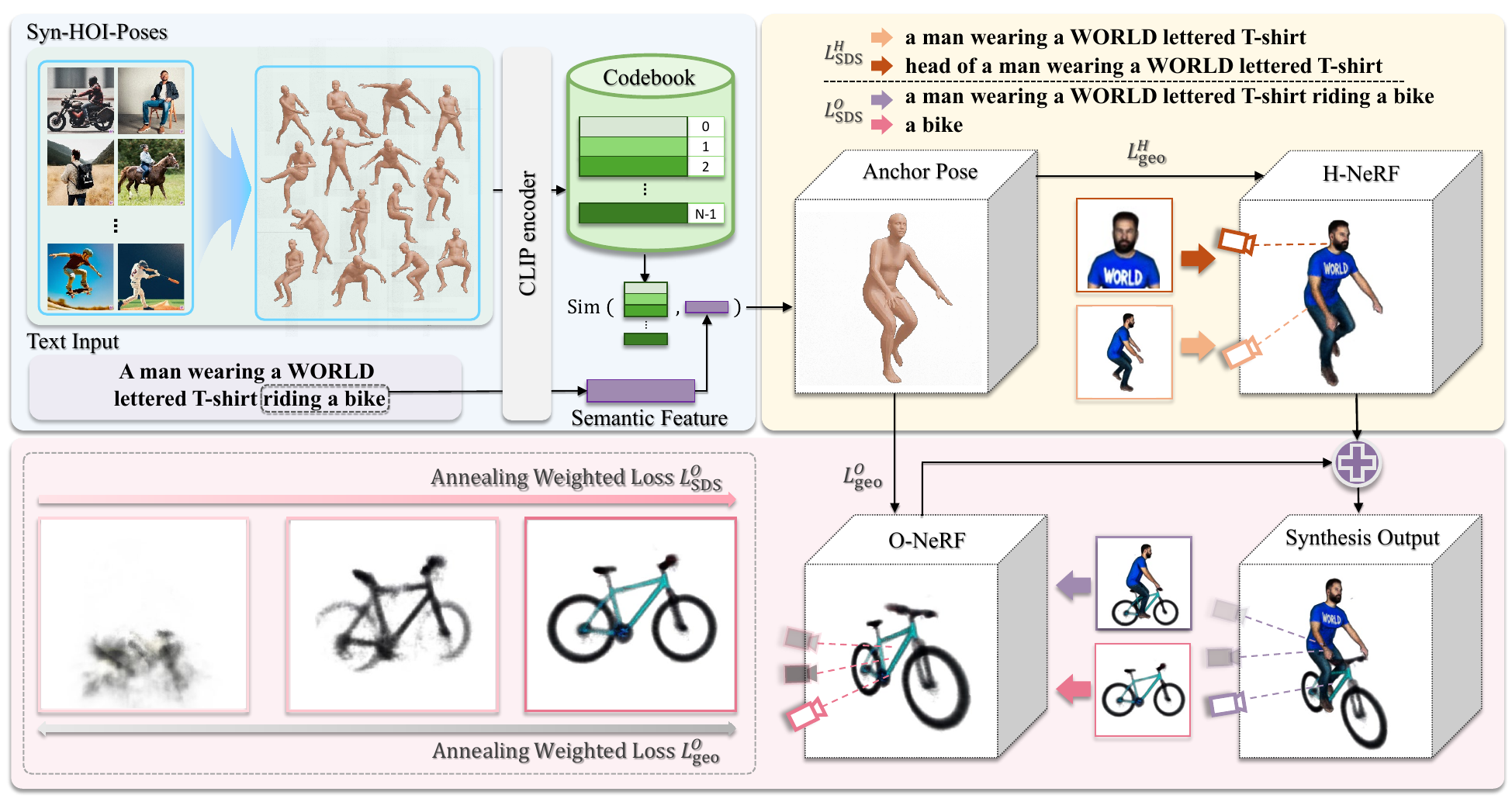}
\caption{InterFusion is a two-stage framework that transforms textual descriptions into detailed 3D human-object interactions, initially synthesizing anchor poses (upper left) and then optimizing the human model (upper right) and object model (bottom) with constraints from estimated pose and textual prompts.}
\label{fig:overview}    
\end{figure*}

As illustrated in Figure~\ref{fig:overview}, our method consists of two primary stages: anchor pose generation and anchor pose guided HOI generation. We first generate an interaction pose based on the input text, termed an anchored pose. This pose then serves as a geometric constraint, guiding the subsequent generation of detailed HOI. In the second phase, the human and object models are optimized separately and refined simultaneously with a global context, ensuring a cohesive and accurate representation of the interaction as described by the input text.

\subsection{Preliminaries}

\delete{
\subsubsection{NeRF.}
Radiance Fields map a 3D position $\spatialpoint$ to a density value and a color value $(\sigma, c)$. Neural Radiance Field (NeRF) \cite{mildenhall2021nerf} implements the mapping functions as multilayer perceptron (MLP) networks. To render the color of a pixel on the 2D image plane by NeRF, the ray would be cast from the camera center through the pixel. The ordered sampled points $\spatialpoint_{i}$ on the ray are then used to query MLPs for their densities and colors $(\sigma_{i}, c_{i})$, which are then accumulated into the expected color with the volume rendering quadrature in accordance with the optical model given by Max \cite{max1995optical}:
\begin{equation}
\begin{aligned}
  & C = \sum_{i} w_{i} {c}_{i}, \\
  & w_{i}=\alpha_{i} \prod_{j=1}^{i-1} (1-\alpha_{j}),
\end{aligned}
\end{equation}
where $\alpha_{i}=1-\exp(-\sigma_{i}\|\spatialpoint_{i+1}-\spatialpoint_{i}\|)$ is the probability of termination at the point $\spatialpoint_{i}$, also indicating occupancy probability.
}

\subsubsection{SDS.}
Score Distillation Sampling (SDS) has been introduced by DreamFusion \cite{poole2022dreamfusion}. While $\renderedimage=\renderer(\modelparams, \camerapose)$, $\renderedimage$ is the 2D image rendered by a differentiable renderer $\renderer$ with model parameters $\modelparams$ (the volumetric renderer and MLPs correspondingly in NeRF), under a desired camera pose $\camerapose$. By injecting the sampled noise $\noise$ into $\renderedimage$ at a time step $\timestep$, the noisy image $\renderedimage_{\timestep}$ is produced. The pre-trained 2D text-to-image diffusion model ${\noisepredictnet}$ provides a denoising network $\hat{\noise}_{\noisepredictnet}(\renderedimage_{\timestep} ; \textembedding, \timestep)$ that predicts the noise $\hat{\noise}$ given the noisy image $\renderedimage_{\timestep}$, time step $\timestep$, and text embedding $\textembedding$. SDS then optimizes the model parameters ${\modelparams}$ by minimizing the difference between the predicted noise and the added noise:
\begin{equation}
\nabla_{\modelparams} \mathcal{L}_{\mathrm{SDS}}(\noisepredictnet, \renderedimage) = \mathbb{E}_{\timestep, \noise} [w(\timestep) (\hat{\noise}_{\noisepredictnet}(\renderedimage_{\timestep}; \textembedding, \timestep)-\noise) \frac{\partial \renderedimage}{\partial \modelparams}],
\end{equation}
where $w(\timestep)$ is the weighting term at the time step $\timestep$.

\subsection{Anchor Pose Generation}
InterFusion starts by generating the anchor pose from the input text $\textprompt^{I}$, which is a non-trivial problem and highly limited by the available pose datasets. Existing datasets (\eg, HumanML3D~\cite{guo2022generating}, BABEL~\cite{punnakkal2021babel}) are mocap based datasets and are typically conducted in a controlled laboratory environment, leading to lacks in action diversity and spatial coverage. 

Recently, text-to-image diffusion models have shown the versatility and effectiveness of generating realistic and diverse images, through the integration of visual and linguistic feature spaces. We thus instead utilize advancements in this technology, specifically models like Stable Diffusion, to create a large-scale dataset of Human-Object Interaction (HOI) images. These images depict a wide range of human interactions. To ensure the diversity of interaction types, we utilize ChatGPT to generate prompts about human daily events or actions forming ``verb-ing a/an/the object''. A total of 235 result prompts are generated, covering most interactions in daily life. After filtering synthesized images without humans, we then estimate 3D human poses using the pre-trained PIXIE~\cite{feng2021collaborative} model, creating a comprehensive Syn-HOI pose dataset consisting of a total 55K 3D pseudo-SMPL poses.

While SMPL model maps pose and shape parameters to a triangulated mesh, to align these poses with our text-to-interaction generation task, we render images from multiple perspectives and use CLIP's image encoder to derive pose feature embeddings. The average feature across these multiple perspective images represents the pose feature and attaches the dataset with pairs of averaged CLIP embedding and pose parameters. We further construct a codebook using K-Means clustering to identify key pose centroids based on the feature embeddings. 2,048 cluster centroids are clustered compositing our pose code-book, where each cluster comprises a subset of poses that represents similar interaction as the key poses of the centroids.

Once the codebook is built, for a given input text, we extract its feature embedding using CLIP's text encoder. This serves as a query to retrieve the most similar poses from the codebook, based on feature similarity. Specifically, given query text $\textprompt^{I}$, top $k$ poses $\bodypose_{k}^{\textprompt^{I}}$ could be matched by pose embeddings $\mathcal{\bodypose}_{E}$, the averaged CLIP embeddings among rendered images from key poses:
\begin{equation}
\bodypose_{k}^{\textprompt^{I}}=\operatorname{TOP}_{k}\left(f_{text}(\textprompt^{I}), \mathcal{\bodypose}_{E}\right)
\end{equation}
where $f_{text}$ is the text encoder of CLIP and $\operatorname{TOP}_{k}\left(X, Y\right)$ returns top $k$ poses with the top $k$ highest cosine similarities, and we use $k=7$ for suitable poses in our experiments.

We then utilize GPT-4V to select the most precise pose as the final queried key pose. Depending on the requirement, we can instead select the poses sampled in the cluster, corresponding to the key pose, to guide the generation process for diverse results. This approach ensures rich and contextually aligned anchor poses for our text-to-interaction generation task.

To restrict the optimization of geometry and appearance using human structural priors from the acquired pose, we further incorporate COAP~\cite{mihajlovic2022coap}, a neural occupancy representation of the articulated human body based on SMPL parameters~\cite{loper2015smpl}. Given a pose $\bodypose$ and a shape $\bodyshape$, COAP offers an occupancy prediction network ${f}(x; \bodyshape, \bodypose)$ that maps a 3D query point $\spatialpoint$ to an occupancy value, directly indicating whether the spatial point resides within the 3D body.

\subsection{Pose-Guided HOI Generation}
Once get the interaction pose, we further use it along with the input text to guide the generation of a detailed 3D HOI scene. In this phase, the estimated pose serves both as spatial constraints for the scene's geometry and as an anchor that aligns the human and object models. This approach ensures that the human model and the object model can be generated separately, while they are cohesively aligned to form the final 3D HOI scene.

The anchor pose provides specific spatial constraints for each component in the scene. For the human model, it establishes a basic geometric structure, while for the object model, it defines the areas that should remain unoccupied. This clear distinction is essential for rendering the scene both accurately and realistically. Simultaneously, the input text undergoes a complex processing procedure to offer distinct semantic guidance for each component. This is accomplished through various text conditioning techniques applied in SDS with the pre-trained DeepFloyd model~\cite{deepfloyd}. The text is carefully crafted to direct the generation of both the human and object models to ensure all elements are in harmony with the semantic nuances of input descriptions.

Additionally, we introduce a novel camera tracing module to enhance the optimization process under varying text conditions. This module adaptively adjusts the camera pose, focusing on relevant elements within the scene at each optimization stage. This adaptive camera positioning is instrumental in ensuring that each aspect of the scene is optimally rendered according to the text descriptions, resulting in a more dynamic and contextually accurate 3D HOI scene. 

We now provide details of each module in this stage, including the Neural Radiance Field representations for the human model (H-NeRF) and object model (O-NeRF), the camera tracing mechanism, and the guided optimization process.

\subsubsection{H-NeRF.}
We use NeRF to represent the human model, noted as H-NeRF. The generation of H-NeRF is guided by the text description specifying the human style, conducting pose-specific human avatar generation with SDS. To enhance the quality of our renderings, particularly in terms of resolution, we incorporated a specialized optimization process that focuses on the head region of the human avatar. The location of the head for any given pose is determined with COAP~\cite{mihajlovic2022coap}. This allows us to augment the text prompt specifically for the head region, using the notation $^{*}\textit{ the head of }^{*}$, to ensure that the head receives detailed attention during the generation process. The loss function for this optimization process is as follows, designed to balance the fidelity of the head region with the overall pose and style of the human figure.

\begin{equation}
\begin{aligned}
\nabla_{\modelparams^{H}} \mathcal{L}_{\mathrm{SDS}}
& = \mathbb{E}_{\timestep, \noise} [w(\timestep) (\hat{\noise}_{\noisepredictnet}(\renderedimage_{\timestep}^{H}; y^{H}, \timestep)-\noise) \frac{\partial \renderedimage^{H}}{\partial \modelparams^{H}}] \\
& + \mathbb{E}_{\timestep, \noise} [w(\timestep) (\hat{\noise}_{\noisepredictnet}(\renderedimage_{\timestep}^{H,h}; y^{H,h}, \timestep)-\noise) \frac{\partial \renderedimage^{H,h}}{\partial \modelparams^{H}}],
\end{aligned}
\end{equation}

In shaping H-NeRF, our geometric constraint ensures that the human model evolves directly from the anchor pose. Points within the anchor are required to be occupied, establishing a firm base for the model. Meanwhile, points outside the anchor can also be occupied, but with probabilities that decrease as they move away from the anchor's surface. This approach allows for the addition of geometric details over the anchor, ensuring the model aligns with the human style described in the text while maintaining a coherent structure rooted in the anchor pose. The loss function used is as follows:

\begin{equation}
\begin{aligned}
  \mathcal{L}_{\mathrm{geo}}^{H}
  & = {\crossentropy}_{{\spatialpoint}_{i} \in \mathbb{\allpoints}_\mathrm{in}}(\alpha_{i}, f({\spatialpoint}_{i})) \\
  & + {\crossentropy}_{{\spatialpoint}_{j} \in \mathbb{\allpoints}_\mathrm{out}}(\alpha_{j}, f({\spatialpoint}_{j})) (1 - e^{-\frac{\distancetoanchor}{2\hyperparameter^2}}), \\   
\end{aligned}
\end{equation}
where $\distancetoanchor$ represents the point distance from the anchor surface, and $\hyperparameter$ is the hyperparameter to control the extent of decaying, similar as in~\cite{metzer2022latent}.

\subsubsection{O-NeRF.}

We also employ NeRF to model the object component, referred to as O-NeRF. This model is designed to interact seamlessly with the human model and to embody the desired style as specified by the input text. To guide the generation of both the interaction type and the object style, we utilize SDS with tailored text prompts:
\begin{equation}
\begin{aligned}
    \nabla_{\modelparams^{O}} \mathcal{L}_{\mathrm{SDS}} 
    & = \mathbb{E}_{\timestep, \noise} [w(\timestep) (\hat{\noise}_{\noisepredictnet}(\renderedimage_{\timestep}^{I}; \textembedding^{I}, \timestep)-\noise) \frac{\partial \renderedimage^{I}}{\partial \modelparams^{O}}] \\
    & + \mathbb{E}_{\timestep, \noise} [w(\timestep) (\hat{\noise}_{\noisepredictnet}(\renderedimage_{\timestep}^{O}; \textembedding^{O}, \timestep)-\noise) \frac{\partial \renderedimage^{O}}{\partial \modelparams^{O}}],
\end{aligned}
\end{equation}

For the interaction scene $x^{I}$ generation, we use an alpha-composited rendering to integrate H-NeRF and O-NeRF. This method calculates an alpha value from the density at each point, determining its contribution to the scene's color. Higher alpha values indicate a greater influence on the rendering, allowing for a nuanced and realistic integration of the human and object models in the final interaction image:
\begin{equation}
\begin{aligned}
  & \renderedimage^{I}  = \sum_{i} w_{i}^{I} {c}_{i}^{I}, 
  w_{i}^{I}=\alpha_{i}^{I} \prod_{j=1}^{i-1} (1-\alpha_{j}^{I}), \\
  & {c}_{i}^{I} = \frac{\alpha_{i}^{H}}{\alpha_{i}^{H}+\alpha_{i}^{O}} c_{i}^{H}+\frac{\alpha_{i}^{O}}{\alpha_{i}^{H}+\alpha_{i}^{O}} c_{i}^{I}. \\
\end{aligned}
\end{equation}

In our NeRF-based approach, alpha values, capped at 1, are derived from density for composite rendering. This setup allows semantic guidance gradients, conditioned by the interaction image, to optimize the density and color of both the human and object models. However, we noted a tendency for the model to prioritize object generation at the expense of the human model's quality. To counteract this, we implemented gradient truncation towards the human model to maintain a balanced optimization between the two components. 

For O-NeRF's geometric constraints, we aim to prevent occupancy of points within the anchor pose, denoted as ${x}_{i} \in \mathbb{X}_{in}$, by the object model. Inspired by the physical collision prevention concept, we define a specific loss function that ensures these points remain unoccupied by O-NeRF, effectively preventing overlap between the human and object models in the 3D space:
\begin{equation}
  \mathcal{L}_{\mathrm{geo}}^{O} = {\crossentropy}_{{\spatialpoint}_{i} \in \mathbb{\allpoints}_\mathrm{in}}(\alpha_{i}, 1 - f({\spatialpoint}_{i})) 
\end{equation}
By minimizing the above loss term, model parameters are optimized to enhance the geometric consistency between H-NeRF and O-NeRF.

\subsubsection{Camera tracing.}
To enhance the generation of O-NeRF, we introduce a dynamic camera tracing module within the SDS framework. This module automatically adjusts the camera pose to focus on the target's center, either the entire interaction scene or just the object during object generation. The camera aligns with the average position of voxels with an occupancy probability above 0.5, ensuring a consistent focus on the most significant parts of the scene or object for optimal detail capture and realism. 

\subsubsection{Optimization.}
The total loss is formulated as follows:
\begin{equation}
\begin{aligned}
  \mathcal{L} = \mathcal{L}_{\mathrm{SDS}}^{H} + \lambda_{1} \mathcal{L}_{\mathrm{SDS}}^{O} + 
  \mathcal{L}_{\mathrm{geo}}^{H} + \lambda_{2} \mathcal{L}_{\mathrm{geo}}^{O} + \lambda_{3} {L}_\mathrm{{reg}},
\end{aligned}
\end{equation}
where $\lambda_{1}$, $\lambda_{2}$ and $\lambda_{3}$ are the corresponding loss weights. Figure~\ref{fig:overview} (left bottom) shows several intermediate states of the object model during the generation steps and we can see that the object is gradually generated with fine details. Weight annealing is adopted during the guided optimization process. We leave the details in the supplement.

\section{Experiments}

Figure~\ref{fig:result} (left) shows results with various interaction poses, showcasing the strength of our method. Our method can support diverse interaction poses within a single type. Meanwhile, with interaction type and pose fixed, our method can further generate more numerous results under different human styles or object styles, as shown in Figure~\ref{fig:result} (right). Moreover, our InterFusion also supports controllable text-conditioned editing, providing users more control over the already generated 3D models, which are presented in the supplementary materials.

\begin{figure*}[t]
    \centering
    \includegraphics[width=\textwidth]{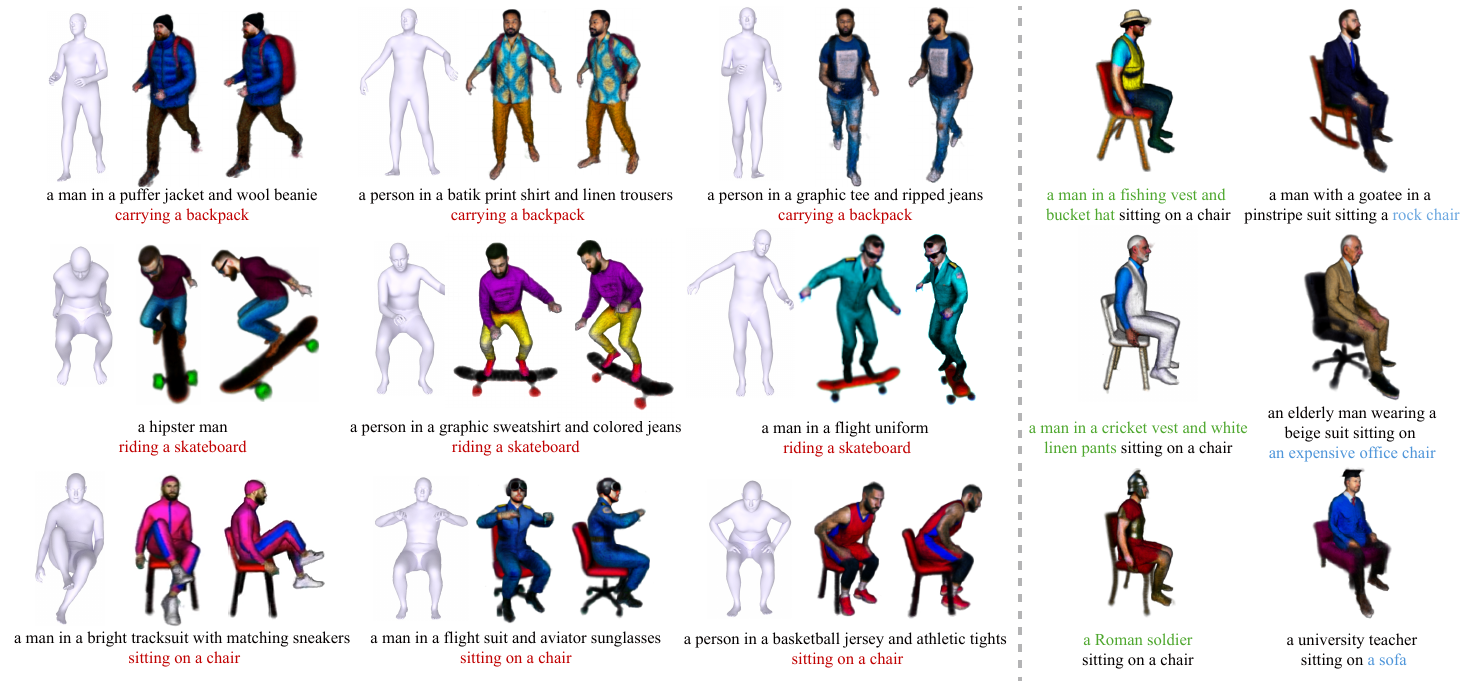}
\caption{More results generated by InterFusion. Diverse integration poses are supported. Numerous human and object styles are also supported.}
\label{fig:result}
\end{figure*}

We both qualitatively and quantitatively evaluate our method against alternative baseline methods. To verify the effectiveness of individual components in our method, we conduct ablation studies. Furthermore, we discuss the application potential, limitations and future work of our InterFusion. The implementation details, more evaluations, and further discussions are presented in the supplementary materials.

\subsubsection{Evaluation prompts and criteria.}

Similar as the first stage for generating prompts about various interaction types, we also use ChatGPT to randomly generate prompts about human styles and collect the text prompts for evaluation. We select 61 distinct and diverse prompts, ensuring the classes of Human (\eg, a policewoman, a teenager) with various styles (\eg, in a graphic tee and ripped jeans), Objects (\eg, a motorcycle, a guitar) and Interactions (\eg, riding, playing) are reasonable and evenly distributed.  In total, we have 13 different types of interactions, covering contact areas across the whole body, which demonstrates the effectiveness of our method on diverse types of interactions compared to baseline methods. 
We calculate the CLIP scores between the input text prompts and different views of generated 3D human-object interactions, and then compare the means based on different evaluation prompts. The CLIP score measures the similarity between a prompt text for an image and the actual content of the image. We also provide an assessment using GPT-4V for selection, named GPT-4V select, which evaluates the completeness of objects and correctness of physical interactions across multiview rendered images. We refer to the supplementary material for assessment details and additional evaluation criteria.

\subsubsection{Baseline approaches.}

To the best of our knowledge, the proposed method is among the first to generate 3D human-object interactions based on text inputs in a zero-shot manner.
We thus compare our method to alternative text-to-3D methods, including DreamFusion~\cite{poole2022dreamfusion}, Magic3D~\cite{lin2022magic3d}, and TextMesh~\cite{tsalicoglou2023textmesh}. Additional comparisons with MVDream~\cite{shi2023mvdream} and ProlificDreamer~\cite{wang2024prolificdreamer} are presented in the supplementary material. To justify our key idea of using human pose to guide the generation, we design an object-centric baseline (Ours-OC) as a variant of our method. Instead of using human pose as the prior, we retrieve an object with the given semantic category from ShapeNet dataset~\cite{chang2015shapenet} and use it as geometry constraints to guide both human and object generation.
As official implementations are unavailable for some of these baselines, we use the third-party re-implementations provided by threestudio ~\cite{threestudio2023} for a fair comparison. Note that all re-implementations use multi-resolution hash-grid ~\cite{muller2022instant} for 3D representation and DeepFloyd~\cite{deepfloyd} for guidance. 

\subsection{Comparison Results}

\subsubsection{Qualitative evaluations.}

\begin{figure}
    \centering
    \includegraphics[width=\linewidth]{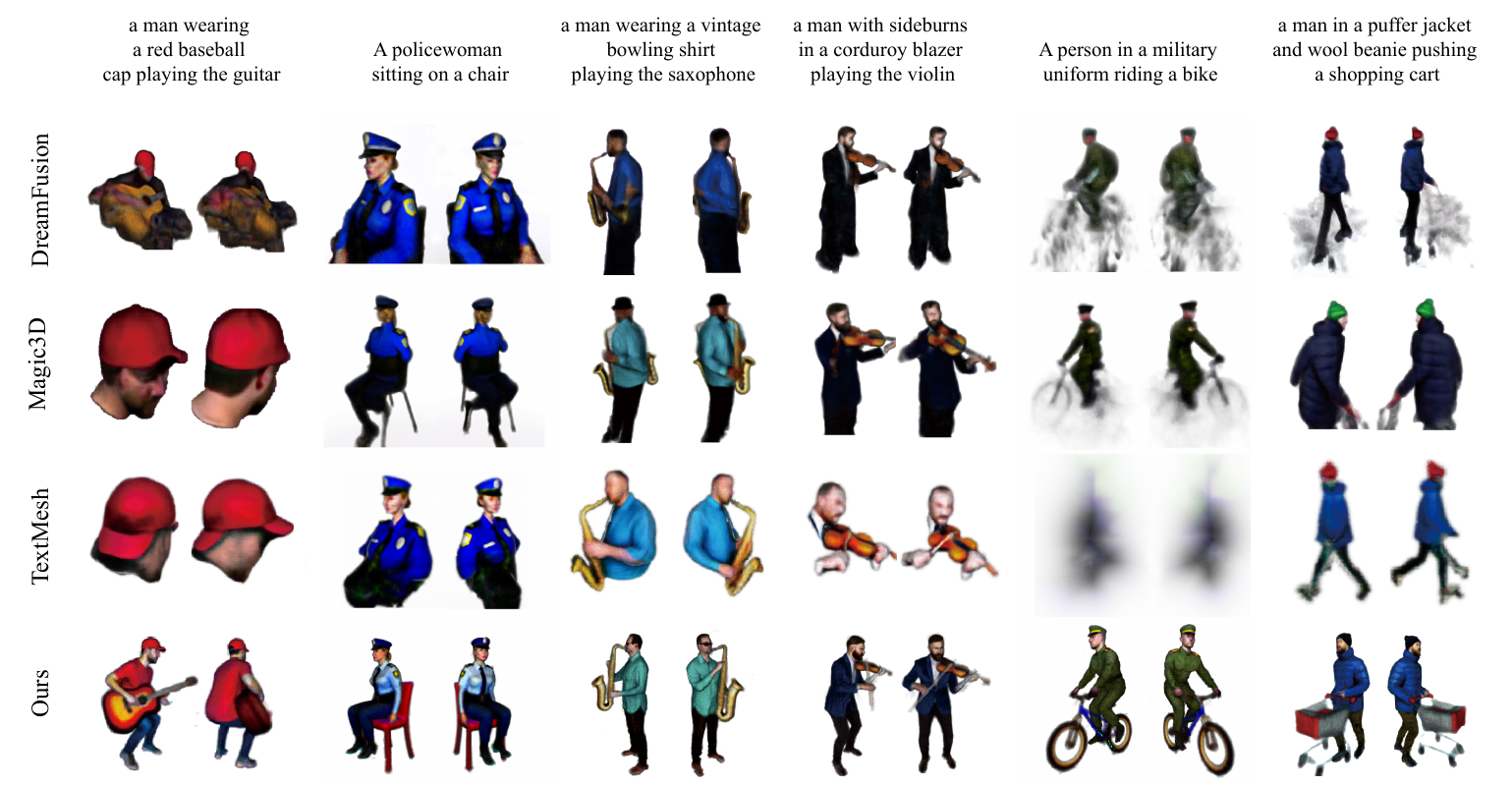}
\caption{Qualitative comparison results with baselines. InterFuion generates more stable and higher-quality results and is more consistent with input interaction descriptions.
}
\label{fig:qualitative}
\end{figure}

Some representative visual comparisons are shown in Figure~\ref{fig:qualitative}, and additional comparisons including those with our object-centric baseline are presented in the supplementary material. The overall results of the baselines reveal common uncertainties in both geometry and appearance attributed to the confusion introduced by multi-concept guidance. Specifically, baseline models may lean towards one specific target, as seen in the example of ``a man wearing a red baseball cap'' with only the red baseball cap generated, the example of ``a policewoman'' with only the upper half of the human body generated, or the example of ``a man in a puffer jacket and wool beanie'' where the shopping cart is failed to be generated. Even with a relatively uniform attention distribution, the baseline model may encounter challenges in producing complete results, as demonstrated in the saxophone and violin examples (the 3rd and 4th column in Figure~\ref{fig:qualitative}), which exhibit deficiencies such as incomplete human body parts, and mixing of human legs. Moreover, the baseline model sometimes struggles to effectively choose the focus of generation when the input text describes a less common target like “a person in a military uniform”, resulting in poor outcomes.

Our approach overcomes these issues by conducting optimization in an explicit decomposed way and intelligently guiding attention from SDS jointly in spatial and semantic aspects. Therefore, our method achieves more stable and higher-quality 3D results under multiple-concept guidance.

\begin{table}[t]
  \centering
    \caption{ Quantitative evaluation of CLIP score and GPT-4V select. }
  \resizebox{0.95\textwidth}{!}{
  \begin{tabular}{l||ccccc}
    \toprule
    Method &DreamFusion~\cite{poole2022dreamfusion} & Magic3D~\cite{lin2022magic3d} & TextMesh~\cite{tsalicoglou2023textmesh} & Ours-OC & Ours-HC\\
    \hline
    CLIP score & 0.3027 & 0.3179 & 0.2761 & 0.3203 & \cellcolor{gray}\textbf{0.3308}\\
    \hline
    GPT-4V select($\%$) & 8.20 & 11.48 & 1.64 & 13.11 & \cellcolor{gray}\textbf{65.57}\\
    \bottomrule
  \end{tabular}}
  \label{tab:example}
\end{table}

\subsubsection{Quantitative evaluations.}

The comparison results are shown in Table~\ref{tab:example}. We can see that our method (Ours-HC) achieves the best performance compared to baselines, showcasing our results exhibit more 3D plausibility with given text prompts. Among the baselines, Ours-OC gets the best performance. Though with object priors, the object-centric approach still does not provide sufficient body priors for human generation, resulting in its inability to achieve complete interaction generation. Magic3D gets the next best performance among the baselines, as it usually generates the whole interaction scene comparing to other two baselines, however, the results are somewhat vague as shown in Figure~\ref{fig:qualitative}, thus there is still a large performance gap comparing to our method. 

The significant enhancement in performance can be attributed to our method's effective concurrent generation of both human and object, along with cohesive interactive information. 

\begin{figure}[tb]
  \centering
  \begin{subfigure}{0.49\linewidth}
        \centering
        \includegraphics[width=0.99\textwidth]{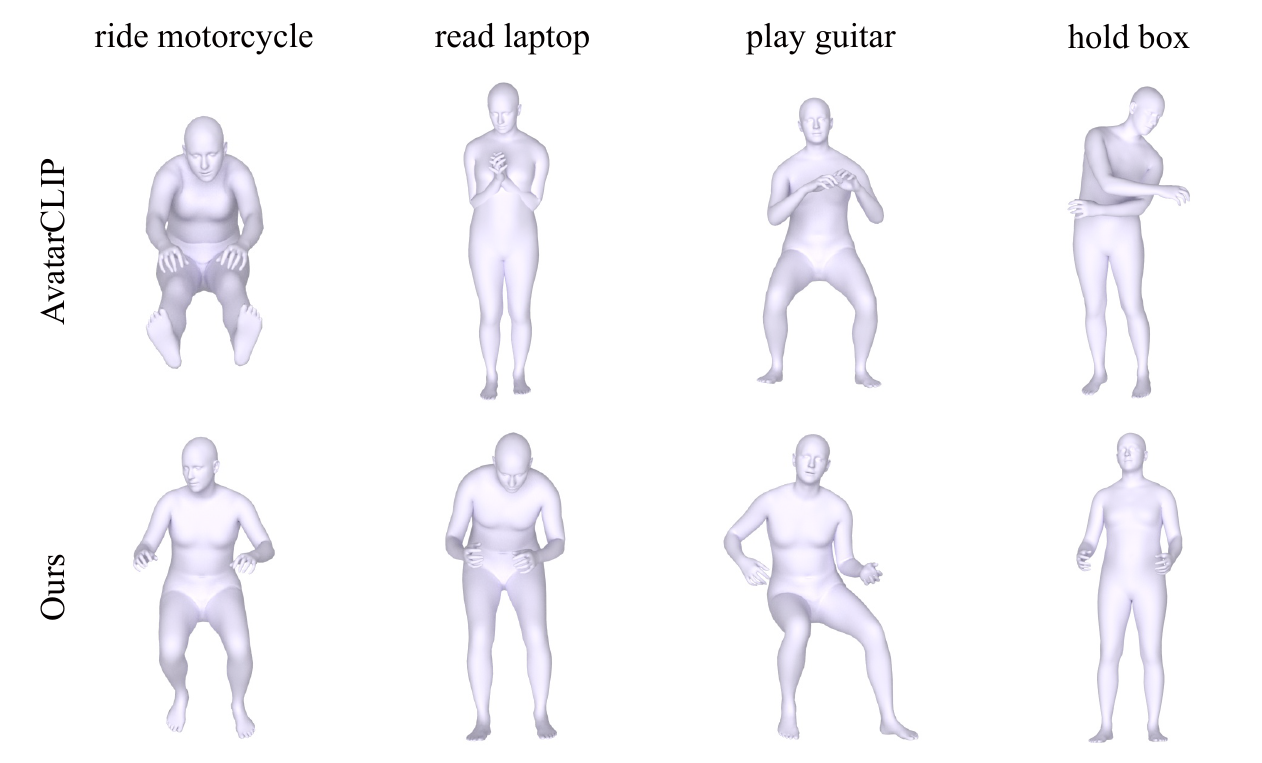}
    \caption{Visual comparison between AvatarCLIP and our pose generation.}
    \label{fig:ablation_left}
  \end{subfigure}
  \hfill
  \begin{subfigure}{0.49\linewidth}
        \centering
        \includegraphics[width=0.99\textwidth]{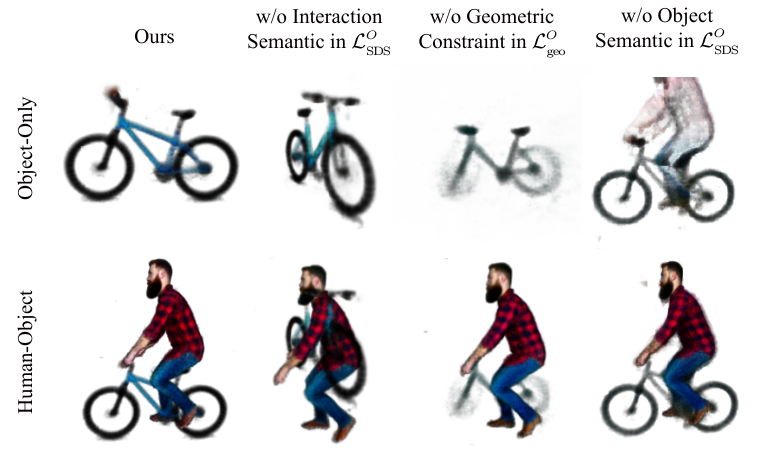}
    \caption{Visual ablation for loss terms during the pose-guided generation process.}
    \label{fig:ablation_right}
  \end{subfigure}
  \caption{Qualitative results of ablation studies.}
\label{fig:ablation}
\end{figure}

\begin{table}[t]
  \centering
    \caption{ Quantitative results of ablation studies.} 
  \begin{tabular}{l||cccc}
    \toprule
    Settings & w/o $\mathcal{L}_{\mathrm{SDS}}^{I}$ & w/o $\mathcal{L}_{\mathrm{SDS}}^{O}$ & w/o $\mathcal{L}_{\mathrm{geo}}^{O}$ & Ours\\
    \hline
    CLIP score & 0.3164 & 0.3293 & 0.3171 & \cellcolor{gray}\textbf{0.3308}\\
    \hline
    GPT-4V select($\%$) & 1.64 & 4.92 & 16.39 & \cellcolor{gray}\textbf{77.05}\\
    \bottomrule
  \end{tabular}

  \label{tab:ablation}
\end{table}

\subsection{Ablation Studies}
 
We conduct several ablation studies both qualitatively and quantitatively to show the importance of our design as well as introduced loss terms. Results are presented in Figure~\ref{fig:ablation_left},  Figure~\ref{fig:ablation_right} and Table~\ref{tab:ablation}. We refer to the supplementary material for additional results and details.

\subsubsection{Syn-HOI-pose.}

The anchor pose obtained at the first stage provides the geometry constraints for our HOI generation. We thus first demonstrate the effectiveness of synthesizing anchoring poses using our pseudo-pose dataset. We compare the pose generated with the existing approach AvatarCLIP \cite{hong2022avatarclip}, which utilizes CLIP only to retrieve the queried pose from the codebook constructed from the mocap dataset AMASS\cite{mahmood2019amass}. Figure~\ref{fig:ablation_left} shows that the proposed pose generation stage enables a better-fitted anchor pose for input interaction types, which demonstrates our pseudo-poses, reconstructed from synthesized images, has the potential for more diverse poses' requirements than existing mocap datasets.

\subsubsection{Semantic guidance.} To better illustrate the efficacy of semantic guidance in our NeRF-based generation, we conduct ablations separately on SDS from object and interaction. As shown in Figure~\ref{fig:ablation_right} and Table~\ref{tab:ablation}, in the absence of SDS from object, semantic consistency for the object is compromised, resulting in noisy outcomes influenced by the human, thus the contact region can not be extracted correctly. SDS from interaction plays the most crucial role in the performance, making results more semantically and visually plausible. The absence of SDS from interaction leads to generated objects being confined to the remaining space, but without interaction with the human.

\subsubsection{Geometric constraint.} Relying solely on semantic guidance is inadequate for achieving the final objectives, as the object should be generated in spaces outside the human body, with its generation targets not positioned at the origin. As illustrated in Figure~\ref{fig:ablation_right} and Table~\ref{tab:ablation}, depending only on semantic guidance may yield results that appear spatially conflict with the human body. Without the spatial constraint, the object generation at the origin would additionally have a conflict with the interaction objective, thus resulting in degenerated objects and final interactions.

\section{Conclusion}

In conclusion, our work presents InterFusion, a novel framework for zero-shot 3D human-object interaction generation. InterFusion tackles the challenges of limited 3D interaction data and the complexity of generating multiple concepts simultaneously. Our two-stage approach, which synthesizes 3D interaction poses from text and then uses these poses as geometric anchors for detailed HOI scene generation, has demonstrated significant improvements over existing methods.

\paragraph{Acknowledgements.} We thank Zejia Su and Yunfan Ye for helpful discussions and paper proofreading. We also thank the anonymous reviewers for their valuable comments. This project was supported by the NSFC (62325211, 62132021, 62322207), Guangdong Natural Science Foundation (2021B1515020085), Shenzhen Science and Technology Program (RCYX20210609103121030), and the Major Program of Xiangjiang Laboratory (23XJ01009).

%

\bibliographystyle{splncs04}
\bibliography{main}

\setcounter{footnote}{0}
\renewcommand{\thefootnote}{\roman{footnote}}
\clearpage
\centerline{\Large{\textbf{Supplementary Materials of InterFusion}}}
\section*{Outline}
In this work, we present InterFusion\footnote{Our code would be accessible at \url{https://github.com/sisidai/InterFusion}.}, a novel zero-shot text-driven 3D human object interaction generation method. We now provide supplementary details in this document, which is arranged as follows:

(1) Sec. A illustrates the implementation details about the methods; 

(2) Sec. B conducts more experiments to verify the superiority of InterFusion;

(3) Sec. C discusses the application potential, limitations and future work.

We also encourage readers to watch our supplementary videos on the project page, which provide more visual representations and perspectives to showcase the 3D properties of our generated human-object interactions.

\appendix

\section{Implementation Details}

We implement InterFusion with threestudio~\cite{threestudio2023}. Specifically, we leverage the multi-resolution hash-grid implementation of implicit volumes in threestudio, along with a Multi-Layer Perceptron (MLP) for predicting density and color values.

\subsubsection{Shading.}
We adopt Lambertian shading with randomly sampled point light during training.
We consider three types of shading, including albedo, diffuse and textureless. 
During training, the shading types of H-NeRF and O-NeRF are enforced to be same for better convergence. 

\subsubsection{Prompting.}
We use one prefix and two suffixes in prompting.
We empirically use the prefix ``a photo of'' to enhance optimization. Additionally, we use the first suffix ``8K, HD'' to improve the resolution and quality. The second suffix is view-dependent and based on the camera location sampled randomly, similar to that in~\cite{poole2022dreamfusion}. Specifically, this view-dependent suffix is set to ''overhead view'' at elevation angles above 60$^{\circ}$. For elevation angles below 60$^{\circ}$, the corresponding text embedding is a weighted interpolation of text embeddings attached with suffixes ``front view'', ``side view'', and ``back view'', where weights are dependent on the azimuth angle. 

\subsubsection{Regularizations.}
Similar to~\cite{poole2022dreamfusion}, several regularization terms are incorporated to enhance the optimization of H-NeRF and O-NeRF, constituting ${L}_\mathrm{{reg}}$. We employ the orientation loss from Ref-NeRF~\cite{verbin2022ref} to encourage normal vectors, that of points along the ray when they are visible, to be forward-facing but not backward-facing to the camera:
\begin{equation}
    \mathcal{L}_{orient} = \sum_{i} \mathrm{stopgrad} (w_{i}) \mathrm{max}(\boldsymbol{n_{i}} \cdot \boldsymbol{v}, 0)^2.
\end{equation}
To encourage the separation from the background and discourage unnecessary floating in empty space, there is also a regularization on the opacity (accumulated the alpha value along each ray):
\begin{equation}
    \mathcal{L}_{opacity} = \sqrt{(\sum_{i} w_{i})^2 + 0.01}.
\end{equation}

\subsubsection{Optimization.}
Recall that our total loss for optimization is:
\begin{equation}
\begin{aligned}
  \mathcal{L} = \mathcal{L}_{\mathrm{SDS}}^{H} + \lambda_{1} \mathcal{L}_{\mathrm{SDS}}^{O} + 
  \mathcal{L}_{\mathrm{geo}}^{H} + \lambda_{2} \mathcal{L}_{\mathrm{geo}}^{O} + \lambda_{3} {L}_\mathrm{{reg}}.
\end{aligned}
\end{equation}
$\lambda_{1}$, $\lambda_{2}$ and $\lambda_{3}$ are the corresponding loss weights, and we adopt weight annealing for them during the optimization process. Specifically, over a total of 10,000 iterations, the weight $\lambda_{1}$ linearly increases from 0 to 1, adding 0.1 every 1,000 iterations. At the outset, the SDS guidance of interaction plays a crucial role initially, providing a good initialization for the object. As the optimization progresses, confidence in the density of the object increases. The weight $\lambda_{1}$ continuously augments, ensuring that the generated components align with the semantic context of the object. As for the weight $\lambda_{2}$, it is empirically set to 0.001 during the initial and final 1,000 iterations, 0.01 during iterations 1,000-2,000 and 8,000-9,000, and 0.1 for the remaining iterations in between. As this weight corresponds to the anchor pose occupancy penalty for the object model, starting with a small value ensures the generation of well-initialized objects from the anchor. Adopting a larger value gradually aids in eliminating redundant human information introduced during initialization, coupled with the SDS guidance from the object. The subsequent decrease in value encourages the final object to contact the human sufficiently, thus aligning more closely with the semantic context of the interaction. The weight $\lambda_{3}$ for the regularization term is constant throughout the optimization process.

\subsubsection{Training details.}
During training, images are rendered under randomly sampled camera views at the resolution of $64\times64$. We use DeepFloyd\footnote{\url{https://github.com/deep-floyd/IF}}, a pre-trained diffusion model, with time steps from $t\sim\mathcal{U}(0.02,0.98)$, and set the weighting function of the time step $\omega(t)$ as 1 consistently.
The classifier-free guidance strength is set to 20.
We use Adam optimizer~\cite{kingma2014adam} with a learning rate of 0.01.
For each 3D scene, the optimization is performed on a single Tesla V100 GPU with 10,000 iterations, requiring approximately 1.5 hours.

\section{Experiments}

\subsection{Additional Comparisons}

\subsubsection{Additional qualitative comparisons.}
\begin{figure}
    \centering
    \includegraphics[width=\linewidth]{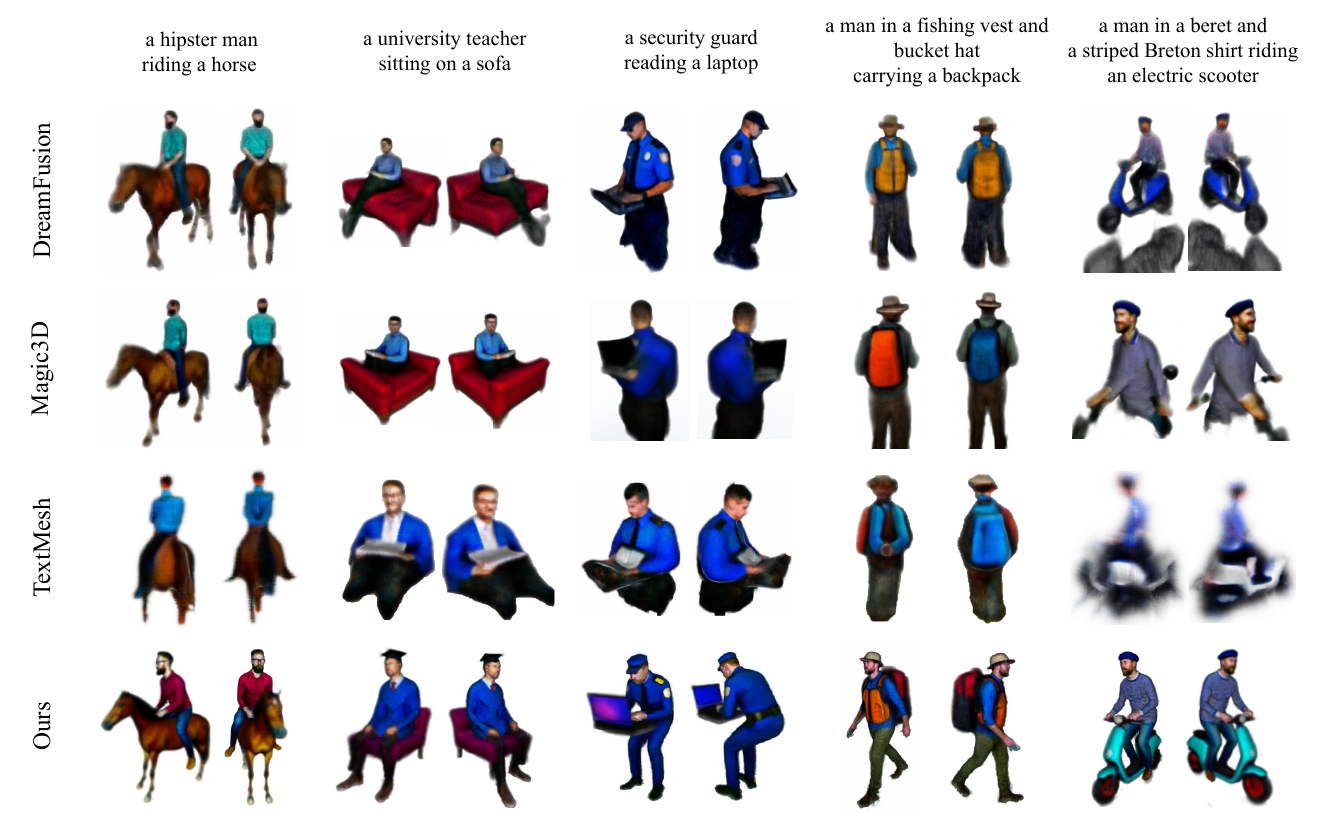}
\caption{Additional qualitative comparison results with baseline methods.}
\label{fig:more_qualitative}
\end{figure}

We have presented qualitative comparisons with several baseline methods, including DreamFusion~\cite{poole2022dreamfusion}, Magic3D~\cite{lin2022magic3d}, and TextMesh~\cite{tsalicoglou2023textmesh}. Qualitative comparisons of additional interaction types with them are shown in Figure~\ref{fig:more_qualitative}. For fairness, the inputs of baselines are also prompted with the same prefix and suffixed as ours. Note that there are two stages in Magic3D: the first NeRF-based~\cite{mildenhall2021nerf} stage as a coarse stage, and the second DMTet-based~\cite{shen2021deep} stage as a refinement stage for higher quality results. We compare our method with its first NeRF-based stage, as ours can be also integrated with a refinement stage. 

\begin{figure}
    \centering
    \includegraphics[width=\linewidth]{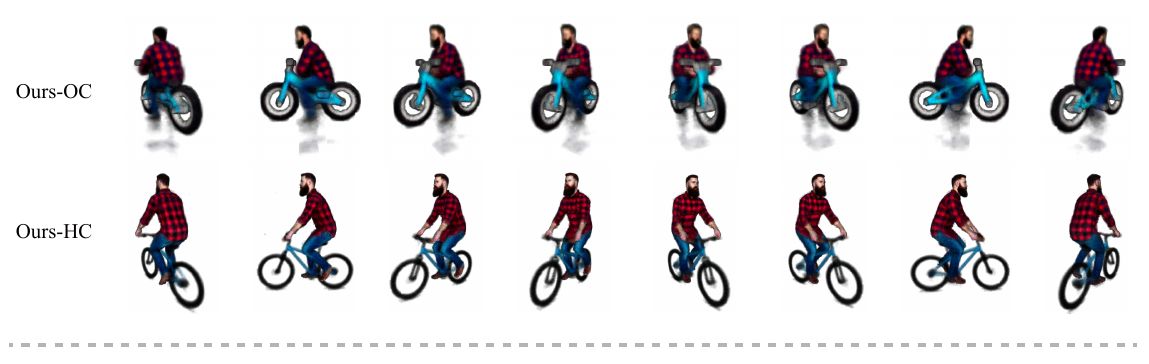}
    \includegraphics[width=\linewidth]{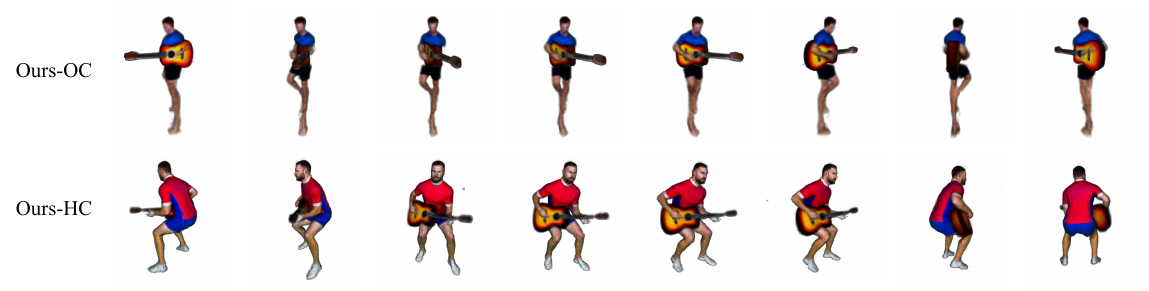}
\caption{Comparison between the object-centric baseline (Ours-OC) and InterFusion (Ours-HC) across multiple views, given the text prompt "a man with a full beard wearing a flannel shirt riding a bike" (top) and "a man in a rugby jersey and cotton shorts playing the guitar" (bottom).
}
\label{fig:oc_qualitative}
\end{figure}

We now provide qualitative comparisons with our designed object-centric baseline (Ours-OC). With object priors, the object-centric baseline more easily generates complete interaction scenes than other baseline methods that start from scratch. Nevertheless, the lack of sufficient human body priors still hampers the ability to achieve complete interaction generation. As seen in Figure~\ref{fig:oc_qualitative}, the object-centric baseline still struggles to generate the full human body, with noticeable absences of body parts involved in interactions and the presence of redundant artifacts.

\begin{table}[t]
  \centering
    \caption{Quantitative comparisons of more baselines and metrics.}
  \resizebox{0.95\textwidth}{!}{
  \begin{tabular}{l||cccccc}
    \toprule
    Method &DreamFusion~\cite{poole2022dreamfusion} & Magic3D~\cite{lin2022magic3d} & TextMesh~\cite{tsalicoglou2023textmesh} & MVDream & ProlificDreamer & Ours\\
    \hline
    R-Precision($\%$) & 68.8 & 73.8 & 47.5 & 77.0 & 67.2 & \cellcolor{gray}\textbf{83.6}\\
    \hline
    $\mathrm{FID}_{\text{CLIP}}$($\%$) & 68.4 & 70.0 & 69.8 & 65.5 & 64.8 & \cellcolor{gray}\textbf{63.7}\\
    \bottomrule
  \end{tabular}}
  \label{tab:add_quantitative}
\end{table}

\begin{figure}
    \centering
    \includegraphics[width=0.8\linewidth]{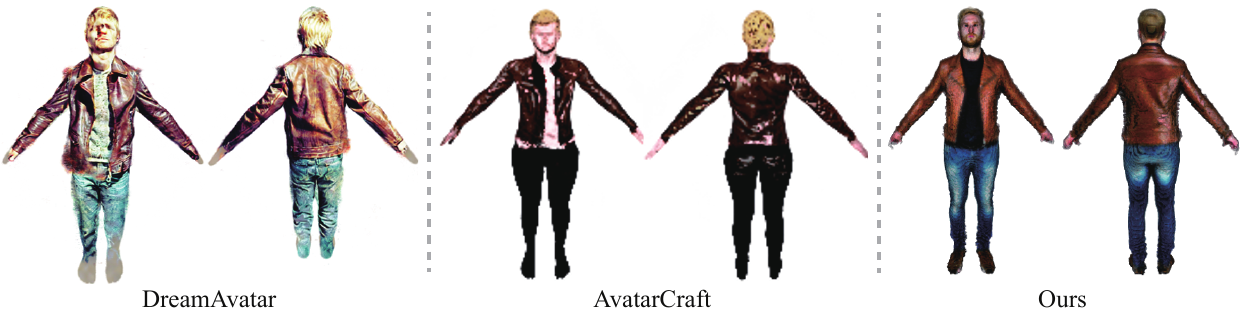}
\caption{Comparisons with recent avatar generation methods, given the text prompt "a man with blond hair wearing a brown leather jacket".
}
\label{fig:avatar_qualitative}
\end{figure}

Moreover, We further compare our method with recent avatar generation methods, including DreamAvatar~\cite{cao2024dreamavatar} and AvatarCraft~\cite{jiang2023avatarcraft}. Visual comparisons are shown in Figure~\ref{fig:avatar_qualitative} and InterFusion achieves competitive quality.

\subsubsection{Additional quantitative comparisons.}
We additionally incorporate CLIP R-Precision and $\mathrm{FID}_{\text{CLIP}}$ into our evaluation metrics, and conduct evaluation to include recent advancements in text-to-3D generation, i.e. MVDream~\cite{shi2023mvdream} and ProlificDreamer~\cite{wang2024prolificdreamer}. The CLIP R-Precision metric~\cite{park2021benchmark}, from the text-to-image generation literature, is the retrieval accuracy with which CLIP~\cite{radford2021learning} retrieves the matching caption among rendered images, evaluates the relevance of the retrieved 3D models to the textual queries. $\mathrm{FID}_{\text{CLIP}}$ assesses the visual fidelity of our generated scenes within the CLIP feature space. These metrics, as shown in Table~\ref{tab:add_quantitative}, underscore our method’s robustness, with our approach outperforming all the methods across all these dimensions.

\subsubsection{Assessment details for GPT-4V selection.}
Though the CLIP score is designed to measure how closely an image aligns with the input text, it falls short in capturing finer details, thus resulting in less pronounced differences in metrics. Inspired by the powerful image understanding capabilities of GPT-4V\footnote{\url{https://chat.openai.com/}}, we further evaluate the performance of baselines and InterFusion over 61 text prompts, using GPT-4V for selection, named GPT-4V select. Specifically, we ask GPT-4V to select one from all generated results with the most 3D justifiability such as full human body, complete object, and correct physical interaction, and then return the index. Note that no in-context examples are given for guidance. Meanwhile, the given order of generated results is randomly shuffled. The answers are summarized in Table~\ref{tab:example}. We also encourage readers to utilize GPT-4V for evaluating the results we have presented, where readers would receive more detailed responses. 

\subsection{Additional Ablations}

\begin{figure}
    \centering
    \includegraphics[width=\linewidth]{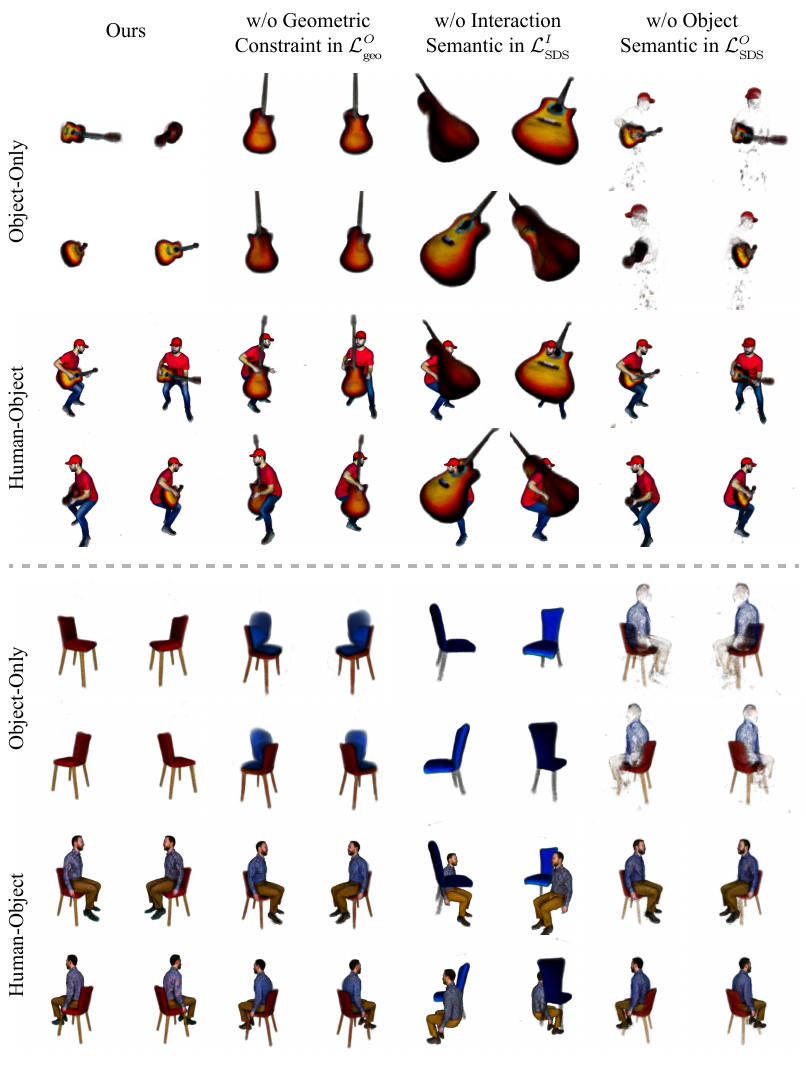}
\caption{ Visual ablations across multiple views for loss terms during the pose-guided generation process, given the text prompt "a man wearing a red baseball cap playing the guitar" (top) and "a person in a paisley print shirt and corduroy pants sitting on a chair" (bottom).
}
\label{fig:more_ablation}
\end{figure}

We provide additional visual examples for loss terms of pose-guided generation in Figure~\ref{fig:more_ablation}, where multiple views of generated results are also provided. As for details of GPT-4V selection, we similarly employ GPT-4V to evaluate the efficiency of loss terms over 61 text prompts. Differently, the object view and the interaction view are both given to GPT-4V in ablations (given object-only in the upper half and human-object in the lower half of the image). We then ask GPT-4V to select one from all generated results with the most 3D justifiability, considering both the complete object and correct physical interaction, and then return the index. No in-context examples are given and the given order of generated results is also randomly shuffled. The answers are summarized in Table~\ref{tab:ablation}. We also recommend readers use GPT-4V for evaluating the results of our ablations.

In general, results generated by our full pipeline are mostly selected, showcasing the collective efficacy of all loss terms. As seen in the 7th and 8th column in Figure~\ref{fig:more_ablation}, results of the absence of SDS from object are mixed with noise from the human body, thus are rarely selected by GPT-4V when considering both the object view and the interaction view. Without the geometric constraint, generations are unstable, resulting in object degeneration and flawed interactions. In some scenarios, the generated object penetrates the human body, with semantically inconsistent interactions (top of the 3rd and 4th column). In rare cases, though the object also intersects, the final interaction remains plausible (bottom of the 3rd and 4th column). Sometimes, such cases would be selected by GPT-4V due to its stochastic nature.



\section{Application Potential, Limitations and Future Work}

\subsection{Application Potential}

Controls for the generated 3D content are challenging and desired. Our InterFusion supports controllable text-conditioned editing, providing users more control over the generated 3D models. Following DreamFusion, we conduct the control by refining the generated 3D model under new given text conditioning. While general text-conditioned editing would modify the geometry and texture in all differing spatial locations, our representation with decomposed human and object enables editing for human-only or object-only within controlled spatial locations. The resulting model preserves the complex spatial relations consistent with the interaction type. 

\begin{figure*}
    \centering
    \includegraphics[width=\textwidth]{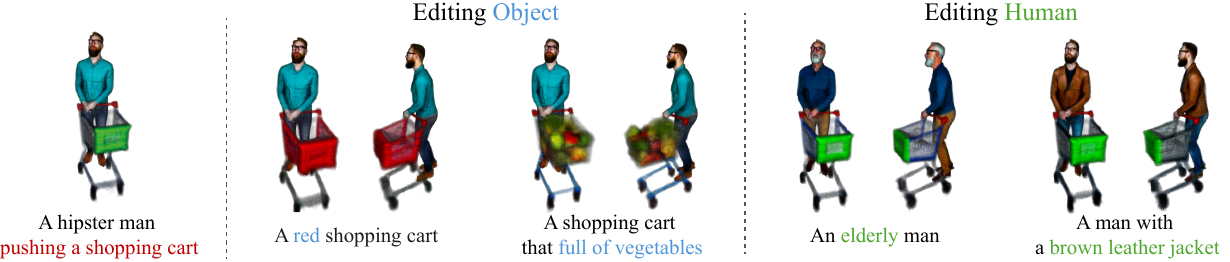}
\caption{InterFuison provides a flexible way for controllable editing of human-object interactions, enabling geometry and texture manipulations for either humans or objects through simple adjustments in the corresponding text prompts. 
}
\label{fig:editing}
\end{figure*}

In Figure~\ref{fig:editing}, we show the model trained with the base prompt for $<$push, shopping cart$>$. Results show that we can refine the human part of the scene model only, e.g. changing the “hipster man” to “elderly hipster man” or “hipster man with a brown leather jacket”. Meanwhile, we can also tune the object part of the scene model only, e.g. changing the “shopping cart” to “red shopping cart” or “shopping cart full of fruits and vegetables”. Both geometry and texture are supported to be edited under new given text conditioning, with the interaction relationship maintained.

\subsection{Limitations and Future Work}

Generating high-fidelity 3D HOI, especially in a zero-shot text-to-3D manner without 3D supervision, is an extremely challenging problem. Our current method primarily focuses on optimizing the global spatial relationship for full-body interactions, thus some inaccuracies in local may still exist, e.g. penetrations at hands. The additional module for hands could be induced in the future.

Our method is also limited by the capabilities of currently used visual language models (VLMs). The progression of VLMs would benefit our method directly. Additionally, we are interested in employing large language models (LLMs) to further enhance our method. Meanwhile, the human-object interaction results generated by our current method are static, we believe extending our framework to incorporate dynamic HOI motions is a good direction for future work.

\end{document}